\documentclass[10pt, a4paper]{article}

\usepackage[final]{lrec2026} 

\usepackage{csvsimple}
\usepackage{caption}
\usepackage{subcaption}
\usepackage{multirow}
\usepackage{booktabs}
\usepackage{float}

\title{HalleluBERT: Let Every Token That Has Meaning Bear Its Weight}

\name{Raphael Schmitt} 

\address{TUM School of Computation, Information and Technology, Technical University of Munich, Germany \\
         Institute of General Practice, Faculty of Medicine and Medical Center, University of Freiburg, Germany \\
         \texttt{raphael.schmitt@uniklinik-freiburg.de}\\
}

\abstract{
Transformer-based models have advanced NLP, yet Hebrew still lacks a RoBERTa encoder that is trained at scale and released in both base and large variants. We present HalleluBERT, a RoBERTa-based encoder family trained from scratch on 49.1~GB of deduplicated Hebrew web text and Wikipedia using a Hebrew-specific byte-level BPE vocabulary. On native Hebrew benchmarks for named entity recognition (BMC, NEMO) and sentiment classification (SMCD), HalleluBERT outperforms monolingual and multilingual baselines, and yields the highest unweighted mean score across the three benchmarks. We release model weights and tokenizer under the MIT license to support reproducible Hebrew NLP research.
 \\ \newline  }

\begin{document}

\maketitleabstract
\pagestyle{plain}
\thispagestyle{plain}

\section{Introduction}\label{sec:introduction}
The emergence of transformer-based language models such as BERT~\citep{devlin_bert_2019} and RoBERTa~\citep{liu_roberta_2019} has transformed natural language processing (NLP), enabling contextualized word representations that generalize across diverse downstream tasks. While multilingual encoders such as mBERT, mmBERT~\citep{marone2025mmbertmodernmultilingualencoder}, and XLM-RoBERTa~\citep{chan_xlm-roberta_2020} offer broad language coverage, monolingual models trained on large, high-quality corpora consistently outperform their multilingual counterparts on language-specific benchmarks~\citep{chan-etal-2020-germans,martin_camembert_2020,delobelle_robbert_2020,scheible-etal-2024-gottbert,scheibleschmitt2025geistbertbreathinglifegerman}. 

For Hebrew, several BERT-style models have been introduced, but most are limited by corpus size, training budget, or absence of large-variant models. In particular, HeRo~\citep{shalumov2023herorobertalongformerhebrew} brought RoBERTa-style pre-training to Hebrew, but appears to have been trained under comparatively limited compute, and does not fully report the effective batch size or total update steps, leaving open how much additional performance could be unlocked under a more extensive pre-training setup. Further, there is no large-scale Hebrew model available.

To address this gap, we introduce HalleluBERT, a RoBERTa-based encoder family trained from scratch on large-scale, cleaned Hebrew web corpora. We release both base and large variants under the MIT open-source license and benchmark them against existing Hebrew-specific and multilingual models.
Our contributions are as follows:
\begin{itemize}
    \item We trained and released HalleluBERT\textsubscript{base} and HalleluBERT\textsubscript{large}, RoBERTa-style encoders for Hebrew.
    \item We benchmarked HalleluBERT on NER and sentiment classification and provide a combined performance analysis against previous state-of-the-art models.
\end{itemize}

\section{Related Work}
Hebrew NLP has seen a steady progression of transformer encoders, from initial Hebrew-specific BERT models to larger-scale pre-training and RoBERTa-style setups.

HeBERT~\citep{Chriqui_2022} was the first transformer-based language model tailored for Modern Hebrew. Trained on Hebrew Wikipedia and OSCAR~\cite{2022arXiv220106642A}, HeBERT improved over multilingual baselines on supervised tasks such as NER, PoS tagging, and sentiment analysis, and formed the basis for HebEMO, a tool for polarity and emotion recognition. However, its training corpus and vocabulary size were comparatively modest, limiting coverage of Hebrew’s morphology and domain diversity.

AlephBERT~\citep{seker2021alephbertahebrewlargepretrained} scaled pre-training to a larger and more diverse corpus (17.9~GB) by combining OSCAR, Wikipedia, and Twitter data. With a 52k vocabulary, AlephBERT established strong results on the Hebrew NLP pipeline and became a widely used baseline.

AlephBERT-Gimmel~\citep{gueta2022large} explores vocabulary scaling for Hebrew by increasing the WordPiece inventory to 128k items and additionally provides a smaller-capacity variant with fewer layers. Their analysis links larger vocabularies to fewer subword splits and reports gains across several Hebrew benchmarks; in our study, we consider only the base model.

HeRo~\citep{shalumov2023herorobertalongformerhebrew} introduced RoBERTa-style pre-training for Hebrew and released the HeDC4 corpus (47.5~GB), a deduplicated and cleaned dataset derived from OSCAR22 and mC4. Building on this resource, they trained HeRo and proposed LongHeRo, a Hebrew Longformer initialized from an intermediate HeRo checkpoint and continued for two additional epochs. The paper does not fully specify key training details such as the effective batch size or the total number of update steps, which can complicate direct comparisons across training budgets.

An overview of existing Hebrew transformer-based models is provided in Table~\ref{tab:hebrew_models}.


\begin{table*}
\centering
\begin{tabular}{llll}
\hline
\textbf{Model} & \textbf{Architecture} & \textbf{Pre-training Data} & \textbf{Corpus Size} \\
\hline
HeBERT & BERT-base & Wikipedia, OSCAR, Emotion UGC & $\sim$10.5 GB \\
AlephBERT & \multirow{2}{*}{BERT-base} & \multirow{2}{*}{OSCAR, Wikipedia, Twitter} & \multirow{2}{*}{$\sim$17.9 GB} \\
AlephBERT-Gimmel & & & \\
HeRo & RoBERTa-base & HeDC4 (mC4 + OSCAR22) & $\sim$47.5 GB \\
LongHeRo & Longformer & HeRo (continued, 2 epochs) & $\sim$47.5 GB \\
HalleluBERT & RoBERTa-base / large & HeDC4 + Wikipedia & $\sim$49.1 GB \\
\hline
\end{tabular}
\caption{\label{tab:hebrew_models} Overview of Hebrew transformer-based language models, including HalleluBERT.}
\end{table*}

\section{Methods}

\subsection{Training Data}
For pre-training HalleluBERT, we used the HeRo dataset (HeDC4)~\citep{shalumov2023herorobertalongformerhebrew} together with a Hebrew Wikipedia dump. HeDC4 is a deduplicated, language-identified, and quality-filtered Hebrew web corpus constructed from mC4 and OSCAR22. In our setup, the HeDC4 portion amounted to approximately 47.5~GB, and Wikipedia contributed an additional 1.6~GB, yielding a total of about 49.1~GB of pre-training text. To mitigate ordering effects from source-specific crawls, we shuffled the documents before pre-training.

\subsection{Pre-processing}
In line with RoBERTa, HalleluBERT employs byte-pair encoding (BPE)~\citep{radford_language_2019} for subword segmentation, operating directly on raw text without requiring pre-tokenization or external tools such as Moses~\citep{koehn_moses_2007}. Since the GPT-2 tokenizer is optimized for English, we constructed a Hebrew-specific tokenizer instead. Following the strategy applied in GottBERT~\citep{scheible-etal-2024-gottbert}, we trained a byte-level BPE vocabulary on 20~GB of shuffled Hebrew text, drawn from both HeDC4 and Wikipedia. This resulted in a 52k subword inventory adapted to Hebrew’s orthographic and morphological properties. Although we did not separately quantify its impact on compression or downstream performance, prior work in Dutch~\citep{delobelle_robbert_2020} and German~\citep{scheible-etal-2024-gottbert} indicates that language-specific tokenizers can yield improvements in both efficiency and accuracy. In practice, we found that sampling about 20~GB was sufficient for subword statistics to stabilize, while scaling vocabulary training to the entire corpus would primarily increase computational cost without offering substantial gains.

\subsection{Pre-training}
Following the setup of GottBERT, we pre-trained HalleluBERT\textsubscript{base} and HalleluBERT\textsubscript{large} using the fairseq framework~\citep{ott_fairseq_2019} on a 128-core TPUv4 pod~\citep{jouppi_tpu_2023}. Due to limitations of our fairseq setup, we did not employ mixed precision; both models were trained in full precision.

HalleluBERT\textsubscript{base} completed training in approximately 30.2 hours, while HalleluBERT\textsubscript{large} required around 6.0 days. We followed the standard RoBERTa pre-training schedule with 100k update steps, a global batch size of 8k, a 10k-step warmup, and polynomial learning rate decay. The base model used a peak learning rate of 0.0004, and the large model 0.00015. 
Although training was configured for 100k steps, the dataset permitted roughly 61 epochs.

\subsection{Evaluation}
Our evaluation design follows the HeRo benchmark suite~\citep{shalumov2023herorobertalongformerhebrew}, which established NER (BMC and NeMO) and sentiment classification (SMCD) as the primary benchmarks for Modern Hebrew, complemented by a Hebrew QA task. 
In our work, we restrict the focus to NER and sentiment classification, building on the NNI- and Huggingface \texttt{transformers}~\cite{wolf-etal-2020-transformers} based grid-search pipeline\footnote{\url{https://github.com/microsoft/nni}} of \citet{he_word_2025}, which we customized for our experiments.
For each model and task, we performed a small grid search over batch size and learning rate. We selected the best configuration based on validation performance and report the corresponding score on the fixed test set; we used a single fine-tuning run per configuration with a fixed random seed (default value of 42).
To stay efficient, we restricted our grid search to batch sizes \{16, 32\} and learning rates \{5e-6, 7e-6, 1e-5, 2e-5, 5e-5\}, based on the most frequent best-performing values in prior experiments (GottBERT, GeistBERT).  
Training was capped at a maximum of 30 epochs for NER and classification tasks, with early stopping applied using a patience of three epochs.  
All models used a linear learning rate schedule with a warmup phase of 10\% of the total training steps. We computed all experiments on two NVIDIA RTX~3090 GPUs.

\subsubsection{Sentiment Classification}
For sentiment analysis in Modern Hebrew, we adopt the benchmark introduced by \citet{amram-etal-2018-representations}. 
The dataset, released by OmiLab, consists of approximately 12{,}800 user-generated social media comments annotated for sentiment polarity (\emph{positive}, \emph{neutral}, \emph{negative}). 
It is provided in two variants: a token-based and a morpheme-based representation, reflecting the morphological richness of Hebrew. 
In line with prior work, we focus on the token-based variant, which has become the standard for model comparison. 

As noted by \citet{seker2021alephbertahebrewlargepretrained}, the original dataset suffered from data leakage between splits due to duplicate samples. 
To address this, we follow the deduplicated version\footnote{\url{https://github.com/omilab/Neural-Sentiment-Analyzer-for-Modern-Hebrew}}, which contains 8{,}465 unique samples after removing duplicates. 
Following HeRo~\cite{shalumov2023herorobertalongformerhebrew}, we refer to this deduplicated corpus as SMCD (Social Media Comments Deduplicated). 

The corpus is distributed with an official train/test split, where the test portion covers roughly 20\% of the data. 
Following common practice, we retain the official test set unchanged for evaluation and extract a validation set comprising 10\% of the training portion to support model selection.
This results in a final distribution of approximately 72\% train, 8\% validation, and 20\% test data.
We report all performance figures on the official test set to ensure comparability with previous studies, using macro-averaged $F_1$ as our primary evaluation metric.

\subsubsection{Named Entity Recognition}
Named Entity Recognition (NER) is a core NLP task that involves identifying and classifying spans of text referring to entities such as persons, organizations, or locations. Early work on Hebrew NER was pioneered by \citet{naama}.  

A key difficulty for Hebrew is its rich morphology, where surface tokens often bundle multiple morphemes. To address this, Bareket and Tsarfaty introduced the NEMO\(^2\) benchmark~\citep{10.1162/tacl_a_00404}, which provides both token- and morpheme-level annotations and has become the standard resource for Hebrew NER.  

In our experiments, we evaluate HalleluBERT on the BMC dataset (split~1)~\citep{naama} as well as on the token-based variant of NEMO\(^2\)~\citep{10.1162/tacl_a_00404}, enabling comparison with earlier baselines while aligning with the current state of the art.
Following standard practice, we report results in terms of the micro-averaged F1 score.

\section{Results}
Table~\ref{tab:ner_results} summarizes the performance of all evaluated models across NER and sentiment classification benchmarks, including a combined average score.

\begin{table*}[h!tbp]
\begin{center}
\begin{tabular}{lccccc}
    \hline
    \multirow{2}{*}{\bfseries Model} & \multicolumn{3}{c}{\bfseries Named Entity Recognition} & \bfseries Classification & \multirow{2}{*}{\bfseries AVG}\\
    & \bfseries BMC & \bfseries NEMO & \bfseries AVG & \bfseries SMCD & \\
    \hline
    \csvreader[late after line=\\]{all_base.csv}{}%
     {\csvcoli & \csvcolii & \csvcoliii & \csvcoliv & \csvcolv & \csvcolvi}
    \hline
    \csvreader[late after line=\\]{all_large.csv}{}%
     {\csvcoli & \csvcolii & \csvcoliii & \csvcoliv & \csvcolv & \csvcolvi}
    \hline
\end{tabular}
\caption{\label{tab:ner_results}
All results are reported as percentages, based on the official test set and the best score out of 10 hyperparameter configurations (selected by validation performance).  
NER performance is measured by micro-$F_1$ on the BMC and NEMO corpora, with the AVG (NER) column showing their unweighted mean.  
SMCD refers to sentiment classification and is measured by macro-$F_1$.  
The rightmost AVG column reports the overall unweighted mean across BMC, NEMO, and SMCD, providing a single combined performance metric.
Best scores are in bold, second-best are underlined, for base and large models respectively.}
\end{center}
\end{table*}

\paragraph{Named Entity Recognition.}
Among the base models, HalleluBERT\textsubscript{base} achieves the highest micro-$F_1$ on both BMC (93.33) and NEMO (87.06), resulting in the strongest NER average (90.20). 
AlephBERT-Gimmel follows closely, slightly outperforming HalleluBERT\textsubscript{base} on classification but falling behind on NER. 
When comparing base to large models, it is noteworthy that HalleluBERT\textsubscript{base} marginally outperforms HalleluBERT\textsubscript{large} on BMC (93.33 vs.\ 93.23), suggesting that the BMC benchmark may not fully reward larger models, possibly due to its limited size or domain coverage. 
However, across NEMO and the NER AVG column, large models retain a small but consistent advantage, confirming that scaling generally benefits NER performance.

\paragraph{Sentiment Classification.}
For classification, AlephBERT obtains the highest $F_1$ among the base models (83.66), slightly ahead of mmBERT\textsubscript{base}. 
HalleluBERT\textsubscript{base} remains competitive but does not lead this task. 
In the large-model category, HalleluBERT\textsubscript{large} achieves the highest classification score (84.91), surpassing XLM-RoBERTa\textsubscript{large} by over one point, further confirming its robustness across tasks.

\paragraph{Overall Average.}
Considering the final AVG column (BMC+NEMO+SMCD), HalleluBERT\textsubscript{large} attains the highest overall score (88.95), with HalleluBERT\textsubscript{base} ranking second (87.83). 
This highlights that HalleluBERT scales gracefully and delivers strong, balanced performance across both NER and classification, outperforming other baselines even when averaged across tasks.

\section{Discussion}
We were unable to reproduce HeRo as the reported state-of-the-art under our standardized fine-tuning pipeline and hyperparameter search space. In our experiments, AlephBERT-Gimmel\textsubscript{base} ranked second among the base models and outperformed HeRo in both NER and overall averages, which may reflect differences in fine-tuning hyperparameters, preprocessing, or evaluation details. We hypothesize that differences in pre-training compute and training setup contributed to the gap. The HeRo paper reports two 35-day training stages on four NVIDIA GTX 1080 Ti GPUs with an initial learning rate of $1e^{-4}$ and linear decay, but key information needed for a like-for-like comparison (e.g., effective batch size and total update steps) is not fully specified~\citep{shalumov2023herorobertalongformerhebrew}.
Moreover, prior work suggests that downstream performance can be substantially influenced not only by corpus design but also by tokenizer and vocabulary choices; for instance, large-vocabulary encoder variants can be competitive across model families~\citep{scheibleschmitt2025sindbertsailorchartingseas}. Related work has also explored BPE vocabulary-size ablations in Turkish~\citep{Toraman_2023}; however, reported trends are not always directly comparable, as they may depend on the underlying pre-training budget and training depth (e.g., data scale, batch size, and number of update steps).

HalleluBERT\textsubscript{base} achieved the best NER results, even slightly surpassing its large variant on BMC (93.33 vs.\ 93.23), likely due to BMC's limited size and domain coverage. Classification was led by AlephBERT, with mmBERT\textsubscript{base} also performing well, possibly benefitting from multilingual pre-training. Despite this, HalleluBERT\textsubscript{base} remained competitive on classification and achieved the highest combined AVG (87.83). The large variant further improved to 88.95, clearly surpassing XLM-RoBERTa\textsubscript{large} (87.49).

Among multilingual models, XLM-RoBERTa\textsubscript{base} outperformed mmBERT\textsubscript{base} despite having fewer parameters, confirming its efficiency as a strong baseline. HalleluBERT, however, outperformed both multilingual models across all tasks, underscoring the value of monolingual pre-training for Hebrew. This contrasts with \citet{scheible-etal-2024-gottbert}, where conservative pre-training learning rates and TPU-based pre-training, lacking dynamic memory allocation and mixed-precision optimization, likely reduced training efficiency and limited the benefits of scaling~\citep{scheible-etal-2024-gottbert}. Under the same fairseq setup and TPU hardware, HalleluBERT achieved markedly better overall results.

Future work includes experimenting with Whole Word Masking (WWM), broader ablations, and throughput analyses (cf.\ PortBERT~\citep{scheible-schmitt-etal-2025-portbert}). In particular, ablating the byte-level BPE vocabulary size (e.g., 32k vs.\ 52k vs.\ 128k) could clarify the performance--efficiency trade-off in terms of sequence length and throughput. Finally, extending evaluation to additional native Hebrew benchmarks would provide a more comprehensive view of model capabilities.

\section{Conclusion}
We presented HalleluBERT, a RoBERTa-style language model for Modern Hebrew, trained extensively on large-scale Hebrew web and Wikipedia text using high batch sizes and long training schedules. Both base and large variants were released and evaluated on native NER and sentiment classification benchmarks, where HalleluBERT achieved state-of-the-art results, surpassing all monolingual and multilingual baselines and delivering the highest overall average across tasks.

These findings highlight the value of large-batch monolingual pre-training for Hebrew and demonstrate that substantial gains are achievable even with conservative hyperparameter choices.

\section{Limitations}
This work has several limitations. First, we used a fixed random seed and do not report confidence intervals or multi-seed averages. Our evaluation targets core encoder-style benchmarks for Modern Hebrew: NER (BMC, NEMO) and sentiment classification (SMCD). We did not include QA or the LongHeRo long-context variant, nor did we extend evaluation to broader suites such as NLI or paraphrase-style tasks. Although translated GLUE-style datasets exist for Hebrew,\footnote{QQP: \url{https://huggingface.co/datasets/imvladikon/qqp_he}, STSB: \url{https://huggingface.co/datasets/imvladikon/stsb_he}} their non-native nature limits comparability with established Hebrew benchmarks.

Second, although we used deduplicated corpora (HeDC4 and Wikipedia), we did not apply additional filtering or cross-source deduplication. As a result, residual noise, low-quality text, and potential biases may remain in the training data and influence the learned representations.

Third, HalleluBERT was trained exclusively on web-based Hebrew text, without explicit control for dialectal or register variation (e.g., Modern Hebrew vs.\ Biblical Hebrew, formal vs.\ colloquial text). This may limit the model's performance on underrepresented varieties or domain-specific language such as legal or medical text, unless additional fine-tuning is performed.

Fourth, we trained HalleluBERT\textsubscript{large} with a conservative peak learning rate of 0.00015 and did not explore extensive hyperparameter tuning. For both model variants we did not apply Whole Word Masking (WWM), which could potentially yield further gains. Pre-training was also performed without mixed precision, which increased computational cost and limited the feasibility of exploring longer training schedules or larger model configurations.

Fifth, we did not perform a detailed error analysis of the model outputs. Such an analysis could provide valuable insights into systematic weaknesses (e.g., common NER boundary errors, sentiment misclassifications) and guide future model and dataset improvements.

\section{Ethical Considerations}
Like all large-scale language models, HalleluBERT may inherit biases from its training data, which can influence downstream tasks such as classification or decision-making. While deduplication reduces redundancy and noise, it does not remove deeper societal or representational biases. Furthermore, training on large web-based corpora raises privacy concerns, as models may inadvertently retain sensitive information. Responsible deployment is especially important in high-stakes domains like legal, medical, or financial NLP.

Despite optimizations for efficiency, pre-training and evaluating transformer models remain computationally demanding, contributing to energy use and carbon emissions. These environmental costs highlight the need for balancing model performance with sustainable development goals.

\section{Acknowledgments}
First and foremost, the author gives all honor and glory to his Lord and Savior, Jesus Christ, whose grace, strength, and guidance made this work possible.

Further, the author gratefully acknowledges the support of Google’s TensorFlow Research Cloud (TFRC) for providing access to Cloud TPUs, which enabled efficient pre-training of HalleluBERT. The author also thanks Nora Limbourg, the assigned Google Cloud Customer Engineer, for her valuable technical assistance and coordination throughout the project.

\section{Bibliographical References}

\bibliographystyle{lrec2026-natbib}
\bibliography{lrec2026-example}


\appendix

\section{Model Properties}
Table~\ref{tab:params_he} summarizes the vocabulary sizes and parameter counts of the Hebrew and multilingual models considered in our evaluation.  
HeBERT (109M) and AlephBERT (126M) represent the first generation of Hebrew BERT-style encoders.  
The HeRo model (125M) and its extended LongHeRo variant (149M) were trained on the HeDC4 corpus and provide strong baselines, particularly for long-sequence tasks.  
Our HalleluBERT\textsubscript{base} (126M) is comparable in size to AlephBERT and HeRo, while HalleluBERT\textsubscript{large} scales up to 357M parameters.  

For multilingual points of reference, mBERT contains 178M parameters with a WordPiece vocabulary of 119k tokens, while XLM-R\textsubscript{base} and XLM-R\textsubscript{large} contain 278M and 560M parameters, respectively, with 250k-token SentencePiece vocabularies.  
All values were extracted using Huggingface’s \texttt{transformers} library.

\begin{table}[H]
\caption{Vocabulary size and total parameter count for Hebrew transformer-based models.  
Values were extracted using Huggingface’s \texttt{transformers} library.}\label{tab:params_he}
\centering\small
\begin{tabular}{lcc}%
    \hline
    \bfseries Model & \bfseries Vocab Size & \bfseries \#Params \\
    \hline
    \csvreader[late after line = \\]{params.csv}{}%
     {\csvcoli & \csvcolii & \csvcoliii}
    \hline
\end{tabular}
\end{table}

\section{Perplexity}
\begin{figure}[htb]
    \centering
    \begin{subfigure}[b]{0.5\textwidth}
         \centering
        \includegraphics[width=\columnwidth]{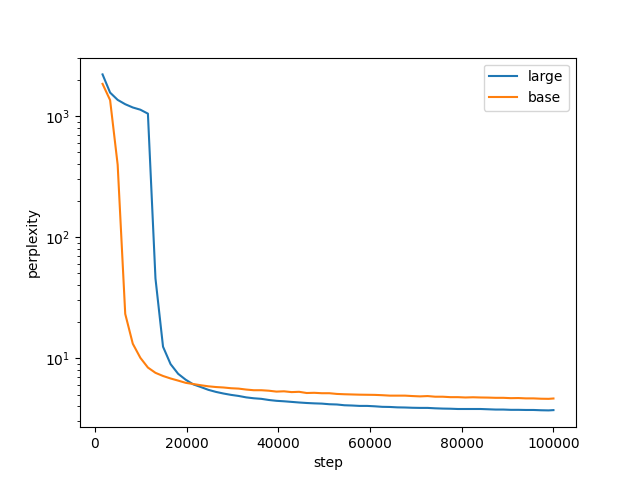}
    \end{subfigure}
    \begin{subfigure}[b]{0.5\textwidth}
         \centering
         \includegraphics[width=\columnwidth]{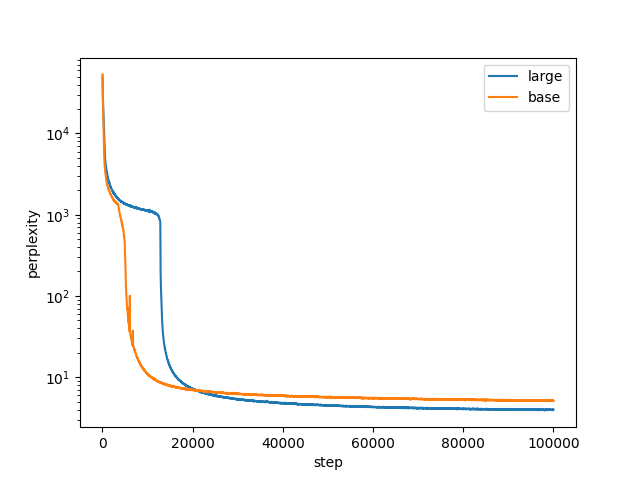}
    \end{subfigure}
    \caption{\label{fig:perplexity}Perplexity of the HalleluBERT models. Top based on a validation at the checkpoints. Bottom based on the validation of each optimization cycle during the training.}
\end{figure}

During pre-training, perplexity was tracked both on the training set (at each optimization step) and on the validation set (after each epoch; see Figure~\ref{fig:perplexity}). Across both model variants, the curves exhibit a plateau phase: this phase is brief for the base models but more pronounced for the large ones. In some cases, short upward spikes occur, which might be misinterpreted as signs of divergence if considered in isolation. The base models typically stabilize after 20k–30k steps, while the large models require slightly longer but also converge by around 40k steps. This overall convergence pattern is mirrored in the validation perplexity, which was evaluated once per epoch.

\section{Parameters}

Table~\ref{tab:best_params} lists the hyperparameters of the best models (selected by validation performance) for each benchmark, supporting reproducibility of our results.  
For transparency, Table~\ref{tab:gpu_time} reports the total computation time per task, showing that all Hebrew downstream experiments together required roughly 65 hours of GPU time (about 2.7~days).


\begin{table}[htb]
    \centering
    \begin{tabular}{lc}
         \hline
         \bfseries Task & \bfseries Computation Time \\
         \hline
         BMC  & 15:43 \\
         NEMO & 27:54 \\
         SMCD & 22:10 \\
         \hline
         Total & 65:47 \\
         \hline
    \end{tabular}
    \caption{Computation time in hours and minutes for the Hebrew downstream tasks, summing to 65 hours and 47 minutes (approximately 2.74 days).}
    \label{tab:gpu_time}
\end{table}

\begin{table*}[!htbp]
\centering
\begin{tabular}{lcccccccc}%
    \hline
    \multirow{2}{*}{\bfseries Model} & 
    \multicolumn{2}{c}{\bfseries BMC} & 
    \multicolumn{2}{c}{\bfseries NEMO} & 
    \multicolumn{2}{c}{\bfseries SMCD} \\
    \cmidrule{2-7}
     & BS & LR & BS & LR & BS & LR
    \\
    \csvreader[late after line=\\]{hyperparams.csv}{}%
     {\csvcoli & \csvcolii & \csvcoliii & \csvcoliv & \csvcolv & \csvcolvi & \csvcolvii}
    \hline
\end{tabular}
\caption{\label{tab:best_params}Hyperparameters of the best downstream task models for each task and pre-trained model. BS refers to batch size, and LR denotes the learning rate.}
\end{table*}

\section{Sequence Length}

As shown in Table~\ref{tab:seq-lengths}, we set maximum input lengths using the 95th percentile of the sequence length distribution for each dataset.
To absorb small pre-processing variations, we added a safety margin and rounded up to the next power-of-two bucket (e.g., 64, 128, 192, 256).
This choice balances efficiency and accuracy by avoiding excessive padding while minimizing truncation.
For the NEMO corpus, although more than 95\% of sequences are shorter than 128 tokens, the maximum observed length is 179.
To cover these rare longer cases, we opted for a maximum length of 192 tokens.
For the SMCD dataset, which exhibits a similar pattern, with a maximum of 1697 tokens but a 95th percentile far below 128, we likewise used 192 tokens as the maximum length.
For the BMC dataset, although sequence lengths never exceed seven tokens, we fixed the maximum length to 64 tokens to match memory-efficient training configurations.

\begin{table*}[htbp]
  \centering
  \begin{tabular}{lccccc}
    \hline
    \textbf{Task} & \textbf{Model} & \textbf{Max Len} & \textbf{Mean Len} & \textbf{95th Pctl.} & \textbf{Seq Len Used} \\
    \hline
    SMCD & HalleluBERT\textsubscript{base}         & 1697 & 30.79 & 96  & \multirow{10}{*}{192} \\
         & HalleluBERT\textsubscript{large}        & 1697 & 30.79 & 96  & \\
         & XLM-RoBERTa\textsubscript{base}               & 2028 & 40.74 & 131 & \\
         & XLM-RoBERTa\textsubscript{large}              & 2028 & 40.74 & 131 & \\
         & mmBERT\textsubscript{small}             & 2606 & 48.82 & 158 & \\
         & mmBERT\textsubscript{base}              & 2606 & 48.82 & 158 & \\
         & HeBERT                                  & 1708 & 31.83 & 99  & \\
         & HeRo                                    & 1680 & 30.47 & 95  & \\
         & AlephBERT                               & 1631 & 30.01 & 94  & \\
         & AlephBERT-Gimmel                        & 1577 & 28.87 & 89  & \\
    \hline
    BMC  & HalleluBERT\textsubscript{base}         & 5 & 3.92 & 4 & \multirow{10}{*}{64} \\
         & HalleluBERT\textsubscript{large}        & 5 & 3.92 & 4 & \\
         & XLM-RoBERTa\textsubscript{base}               & 4 & 3.78 & 4 & \\
         & XLM-RoBERTa\textsubscript{large}              & 4 & 3.78 & 4 & \\
         & mmBERT\textsubscript{small}             & 7 & 6.50 & 7 & \\
         & mmBERT\textsubscript{base}              & 7 & 6.50 & 7 & \\
         & HeBERT                                  & 5 & 3.85 & 4 & \\
         & HeRo                                    & 5 & 3.90 & 4 & \\
         & AlephBERT                               & 4 & 3.82 & 4 & \\
         & AlephBERT-Gimmel                        & 4 & 3.46 & 4 & \\
    \hline
    NEMO & HalleluBERT\textsubscript{base}         & 106 & 29.02 & 53 & \multirow{10}{*}{192} \\
         & HalleluBERT\textsubscript{large}        & 106 & 29.02 & 53 & \\
         & XLM-RoBERTa\textsubscript{base}               & 151 & 39.94 & 76 & \\
         & XLM-RoBERTa\textsubscript{large}              & 151 & 39.94 & 76 & \\
         & mmBERT\textsubscript{small}             & 179 & 48.15 & 91 & \\
         & mmBERT\textsubscript{base}              & 179 & 48.15 & 91 & \\
         & HeBERT                                  & 110 & 28.77 & 54 & \\
         & HeRo                                    & 108 & 28.64 & 52 & \\
         & AlephBERT                               & 102 & 27.40 & 51 & \\
         & AlephBERT-Gimmel                        & 100 & 26.03 & 48 & \\
    \hline
  \end{tabular}
  \caption{Sequence length statistics for all benchmark datasets (SMCD, BMC, NEMO) across evaluated models. Reported are maximum observed length, mean length, 95th percentile, and the sequence length used during training.}
  \label{tab:seq-lengths}
\end{table*}

\end{document}